\documentclass[11pt]{article}

\usepackage[margin=1in]{geometry}

\usepackage[ngerman,british]{babel}
\usepackage{natbib}
\setcitestyle{aysep={}} 

\usepackage{url}
\usepackage{pdflscape}

\usepackage{array}
\usepackage[flushleft]{threeparttable}
\usepackage{rotating,graphicx}
\usepackage{setspace}
\usepackage{color}
\usepackage{mathtools}

\usepackage[flushmargin]{footmisc}
\usepackage{titlesec}
\titlelabel{\thetitle.\ }

\definecolor{CardinalRed}{cmyk}{0,1,0.65,0.34} 
\usepackage[colorlinks,linkcolor=CardinalRed,citecolor=CardinalRed,urlcolor = CardinalRed]{hyperref}
\usepackage[labelsep=period, font=sc, skip=0pt,justification=centering]{caption}

\usepackage{placeins}

\RequirePackage{amsmath}
\RequirePackage{amssymb}
\RequirePackage{amsbsy}
\RequirePackage{amsthm}
\RequirePackage{amsfonts}
\RequirePackage{latexsym}
\RequirePackage{graphicx}

\def\mydetails#1{\thanks{Albert Chiu, PhD Candidate, Department of Political Science, Stanford University. Email: \url{altchiu@stanford.edu}.{#1}}}

\newcommand{\mc}[1]{\mathcal{#1}}

\newcommand{\R}{\mathbb{R}}

\def\E{\mathbb{E}} 

\providecommand{\argmax}{\mathop{\rm argmax}} 

\providecommand{\supp}{\mathop{\rm supp}}

\newcommand{\uniform}{\mathsf{Uni}}

\newcommand{\fisher}[1][]{%
  \ifthenelse{\isempty{#1}}{
    I
  }{%
    I_{#1}
  }
}

\theoremstyle{definition}

\newenvironment{proof-sketch}{\noindent{\bf Sketch of Proof}
  \hspace*{1em}}{\qed\bigskip\\}
\newenvironment{proof-idea}{\noindent{\bf Proof Idea}
  \hspace*{1em}}{\qed\bigskip\\}
\newenvironment{proof-of-lemma}[1][{}]{\noindent{\bf Proof of Lemma {#1}}
  \hspace*{1em}}{\qed\bigskip\\}
\newenvironment{proof-of-proposition}[1][{}]{\noindent{\bf
    Proof of Proposition {#1}}
  \hspace*{1em}}{\qed\bigskip\\}
\newenvironment{proof-of-theorem}[1][{}]{\noindent{\bf Proof of Theorem {#1}}
  \hspace*{1em}}{\qed\bigskip\\}
\newenvironment{inner-proof}{\noindent{\bf Proof}\hspace{1em}}{
  $\bigtriangledown$\medskip\\}
\newenvironment{proof-attempt}{\noindent{\bf Proof Attempt}
  \hspace*{1em}}{\qed\bigskip\\}

\usepackage{algorithm} 
\usepackage{algorithmic}
\usepackage{subcaption}

\newcommand{\itehat}{\hat\tau_{i}}

\newcommand{\complexity}{complexity}
\newcommand{\length}{\text{length}}

\newcommand{\add}{\texttt{ADD}}
\newcommand{\cut}{\texttt{CUT}}
\newcommand{\replace}{\texttt{REPLACE}}
\newcommand{\addc}{\texttt{ADD}$_{cond}$}
\newcommand{\cutc}{\texttt{CUT}$_{cond}$}
\newcommand{\replacec}{\texttt{REPLACE}$_{cond}$}

\title{An Algorithm for Identifying Interpretable Subgroups With Elevated Treatment Effects}
\author{Albert Chiu\mydetails{ I'm grateful to Yiqing Xu, Dominik Rothenhausler, Justin Grimmer, Jens Hainmueller, Emma Brunskill, and participants at the Stanford Causal Science Center Conference on Experimentation for helpful comments and suggestions.}}
\date{\today}

\begin{document}
\onehalfspacing
\maketitle
\abstract{We introduce an algorithm for identifying interpretable subgroups with elevated treatment effects, given an estimate of individual or conditional average treatment effects (CATE).
Subgroups are characterized by ``rule sets''---easy-to-understand statements of the form \texttt{(Condition A AND Condition B) OR (Condition C)}---which can capture high-order interactions while retaining interpretability. 
Our method complements existing approaches for estimating the CATE, which often produce high dimensional and uninterpretable results, by summarizing and extracting critical information from fitted models to aid decision making, policy implementation, and scientific understanding.
We propose an objective function that trades-off subgroup size and effect size, and varying the hyperparameter that controls this trade-off results in a ``frontier'' of Pareto optimal rule sets, none of which dominates the others across all criteria.
Valid inference is achievable through sample splitting.
We demonstrate the utility and limitations of our method using simulated and empirical examples.
}

\doublespacing
\section{Setting}

In causal inference, average treatment effects (ATE) and average treatment effects on the treated (ATT) are the estimands that garner the most interest. 
Even if the effect of a treatment is known to be positive on average, it can vary greatly across individuals; some individuals will benefit, but some may experience no effect, and others may even be hurt. 
A large body of literature on heterogeneous treatment effects (HTE) has developed with the common goal of estimating a conditional average treatment effect (CATE) function $\tau(X)$, which maps individuals' characteristics onto treatment effects \citep{wager-athey_2018_forest, nie2021quasi, kunzel2019metalearners, grimmer2017estimating, imai2013estimating, athey2016recursive}.

This literature often motivates itself either from a scientific understanding perspective (i.e., some theories inherently imply heterogeneity in effects) or as a means to help design treatment regimes. 
In medicine, industry, government, and elsewhere, there are ample applications where learning who benefits the most from a treatment can improve outcomes. 
One practical challenge is that the CATE can be a very high dimensional function that is uninterpretable and of limited direct utility.

This paper proposes a method to supplement existing CATE estimators by summarizing their output and producing understandable descriptions of the CATE that are actionable and from which we can learn.
In particular, we propose an algorithm for learning rule-based treatment assignment rules that are interpretable and capture complex interactions between variables. 
We propose an objective function that trades-off between the size of the treatment group and the size of the treatment effect. By varying a hyperparameter of the objective function, we can adjust the relative importance of each desiderata. Each value of the hyperparameter corresponds to a Pareto optimal solution that is not dominated in both desiderata by any other feasible rule set.
We use simulated and empirical examples to demonstrate the utility and limitations of this method.

\section{Related Work}

\subsection{Policy Learning and Targeted Treatments} 

Our work most directly contributes to the literature on learning treatment assignment ``policies.'' Policies, also referred to as individualized treatment rules or treatment regimes, is a mapping from observed characteristics to a treatment assignment, $\pi:\mc{X}\to [0,1]$.

Precision medicine, which seeks to improve healthcare outcomes through better-targeted treatments, motivates much of the existing work in this area \citep{kosorok2019precision}. 
\citet{qian2011performance} propose using an $l_1$-penalized least squares estimator to select for relevant covariates to include in a treatment rule.
\citet{zhang2012estimating} propose transforming the problem into a weighted classification problem, allowing the use of any classifier that can accommodate weights.
\citet{zhao2012estimating} also take a weighted classification approach, where the weights are proportional to the patient's outcome.
\citep{foster2011subgroup} propose the ``Virtual Twins'' method, whereby they estimate the difference in expected outcome for each patient and a hypothetical patient with identical characteristics but different treatment status, and then this estimate as the outcome in a classification or regression tree.

Proposals have come from economics as well.
\citet{kitagawa2018should} develops the empirical welfare maximization (EWM) method, which selects the policy that maximizes in-sample total welfare over a class of feasible policies. This feasible class is meant to encode constraints, including ethical, legal, budget, and interpretability constraints. 
A closely related proposal by \citet{athey-wager_2021_policy} also seeks to find a total utility maximizing policy from within a limited class, but the authors provide theoretical guarantees under more general conditions.

A key tension in this literature is the one between the interpretability and the richness of the function class used. Proposals often either risk being too simplistic and unable to capture complex dependencies between variables (e.g., linear separation) or still permit overly complex policies that that are not suitable for most applications (e.g., deeply grown trees).
This paper argues that rule sets represent the best of both worlds and can capture complex interactions while remaining interpretable. 
\citet{wang2022causal} also take the approach of using rule sets to describe treatment assignment rules, but they address a more narrow setting where the outcome is binary and modeled using a Bayesian logistic regression. Our proposal is suitable for a much wider range of settings, and we argue that our choice of objective function makes more clear the trade offs between desiderata.
We reserve a full discussion of rule sets and how they compare to other proposals for a later section.

\subsection{Multi-Objective Optimization}
We also make an indirect contribution to the literature multi-objective optimization. A $d$-objective objective function is a mapping from the solution space $\mc{A}$, the data $Z$, and hyperparameters $\alpha\in R^u$ onto $\R^d$, $F:\mc{A}\times Z \times \R^u\to \R^d$. 
A common criteria for solutions to such optimization problems is Pareto efficiency or optimality, which states that no objective can be improved without worsening another objective.

The task of simultaneously optimizing over multiple objectives is difficult, so most proposals ``scalarize'' the objective function, i.e. reduce the dimensionality of the objective function to one.
\citet{czyzzak1998pareto}, for example, propose Pareto simulated annealing (PSA) as a metaheuristic for solving multi-objective combinatorial optimization problems, but they scalarize the multi-objective objective function into a linear weighted sum, $\sum_l w_i f_i(A)$, which is known to be incapable of finding all Pareto optimal solutions when the Pareto front is non-convex \citep[e.g.,][]{miettinen1999nonlinear}. The key insight is that the linear scalarization produces linear level curves, which cannot be tangent to non-convex parts of the Pareto front.

The Chebyshev or hypervolume scalarization is guaranteed to find all Pareto optimal solutions \citep[Theorem 3.4.5. in][]{miettinen1999nonlinear}, but it involves solving a more difficult minimax problem, $\min_A \max_i w_i f(A)-z_i^*$ for some set of convex weights $w_i$ and a reference vector $z^*$. \citet{schultes2021hypervolume} give some geometric intuition for this result: For a well-picked reference point, the level curves of the hypervolume scalarization resemble those of an $l_\infty$ function, which has a sharp corner that can be tangent to any part of a Pareto front of any shape.

Our approach straddles the middle. We use a multiplicative objective function, $f_1(A)^\alpha f_2(A)$ for $\alpha >0$. This is similar to the Cobb-Douglas utility function that is common in economics, $f_1(A)^\alpha f_2(A)^{1-\alpha}$, but we prefer ours for interpretive reasons that we discuss later. Our objective function produces hyperbolic level curves that still cannot reach some solutions but is easier to solve than the hypervolume problem.

\section{Pareto Efficient Rule Sets}


\subsection{Rule Sets and Interpretability}

Rule sets are if-then statements that take the form of ``OR'' statements composed of ``AND'' statements, such as ``IF ($A$ AND $B$) OR (C) THEN D=1.''
In set theoretic terms, they are unions of intersections. Let $C$ be a collection of conditions, where each condition $c\in C$ is a set, and we say that an observation $i$ satisfies a condition $c$ if $i\in c$. The intersection of the conditions in $C$ is a rule $a=\bigcap_{c\in C} c$, and we say that the length of $a$ is the number of conditions in $C$, length$(a)=|C|$. The union of a collection of $L$ rules $\{a_l\}_{l=1,\dots,L}$ is a rule set of size $L$.
We say a rule $a$ or rule set $A$ covers an observation $i$ whenever $i\in a$ or $i\in A$. We may also say an observation $i$ satisfies a rule or a rule set to mean the same thing. 

Importantly, they are interpretable: For any given observation, it is easy to see the reasoning behind a recommendation to be treated or not. 
A large body of work has already argued the importance of interpretability, in particular for high-stakes decision making \citep{rudin2019stop}, when there is incompleteness in our understanding of the problem formulation that has the potential for unintended and unquantified bias \citep{doshi2017towards}, or when there is a mismatch between real-world objectives such as ethics and legality and those encoded in our evaluation metrics \citep{lipton2018mythos}.
In many contexts, such as medicine and criminal justice, ``treatment'' assignment rules satisfy each of these criterion.

Generally, more parsimonious rule sets are more interpretable, and there are two dimensions of parsimony: number of rules in a rule set and length of each rule. 
We define the complexity of a rule set $A$ to be the sum of the lengths of all rules in $A$, $\complexity(A)=\sum_{a\in A}\length(a)$.
To ensure domain-specific interpretability, we limit the length of rules to $L_{max}$ and the complexity of rule sets to $C_{max}$.

\subsection{Comparison With Trees}


The most apt comparison to make is with decision trees, which are a popular existing way of representing treatment rules and have a bijective relationship with rule sets.
Despite this bijection, we argue that rule sets correspond to richer, more interpretable classes of policies than trees: To achieve the same level of descriptiveness as rule sets, we would need to expand the class of trees under consideration to a point where we admit uninterpretable trees as well.

The most natural way of constraining the complexity of a tree is to restrict its depth.
\citet{athey-wager_2021_policy}, for example, set the universe of policies to be decision trees of depth $L=2$ in their examples. 
Although trees of depth $L$ can be represented as rule sets of at most $2^{L-1}$ rules of length $L$, the reverse is not true. 

\begin{figure}[h!]
\caption{Decision Trees}
\label{fg:tree-vs-rs}
\hfill
\begin{subfigure}[t]{0.45\textwidth}
         \includegraphics[width=\textwidth]{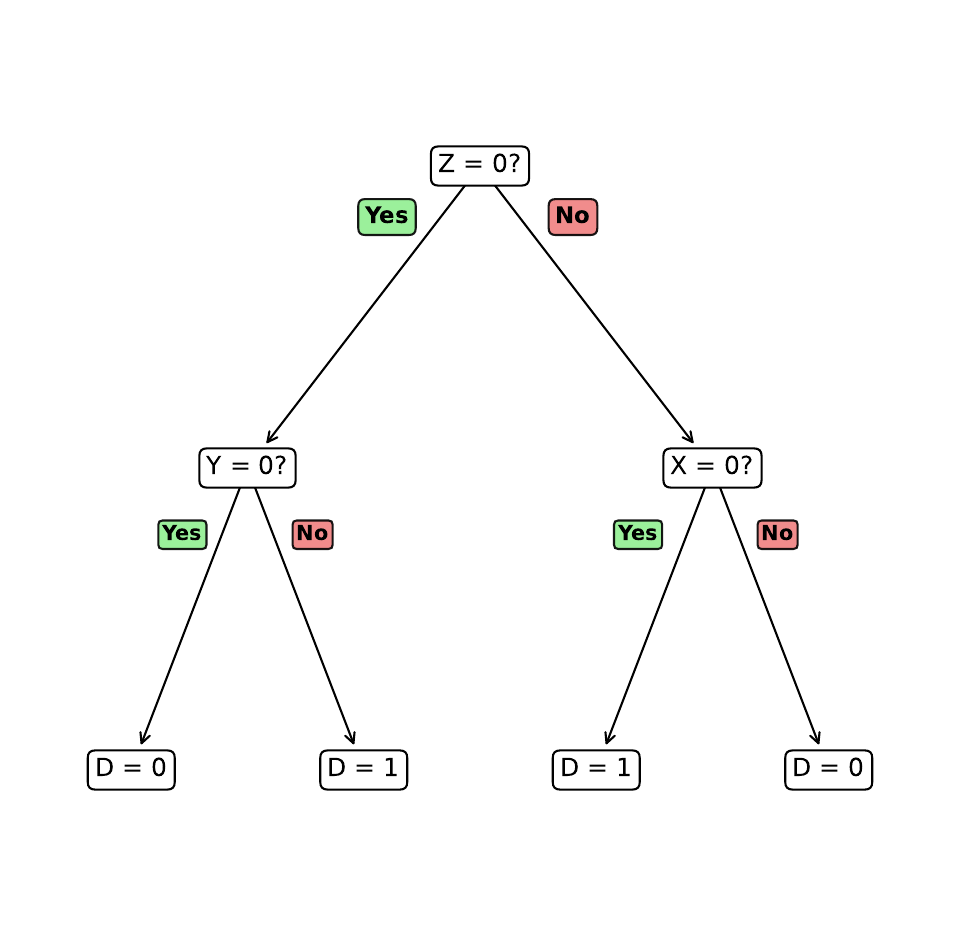}
         \caption{Example Tree}
         \label{fg:tree-eg-depth2}
     \end{subfigure}
     \hfill
     \begin{subfigure}[t]{0.45\textwidth}
         \includegraphics[width=\textwidth]{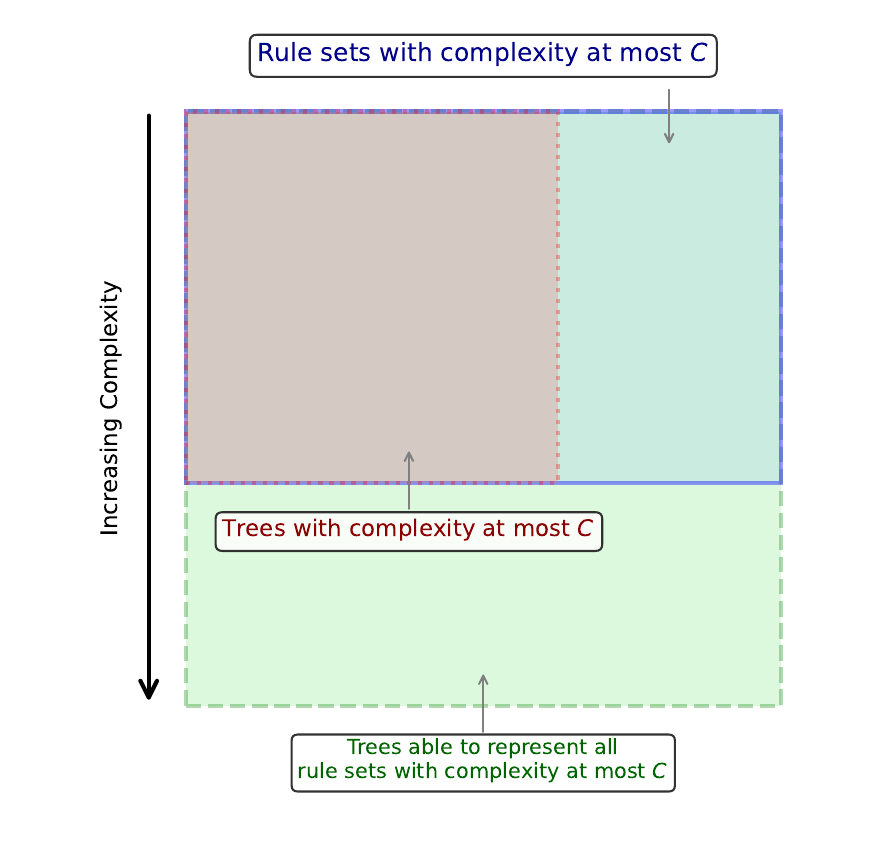}
         \caption{Complexity and Policy Class Size: Rule Sets vs. Depth-Constrained Trees}
         \label{fg:complexity}
     \end{subfigure}
     
     \footnotesize{\textbf{Note:} (left) An example of a decision tree of depth two, and (right) a diagram of the relationship between complexity and the richness of a policy class for trees and rule sets.}
\end{figure}

The key observation here is that any decision tree can be translated into a rule set, but there are restrictions on the type of rule sets that a decision tree of a fixed depth or complexity can represent.
Each path in a decision tree that terminates in a positive classification (in our case, $D=1$) corresponds to a rule, and each node along that path is a condition in that rule. We use the tree in Figure~\ref{fg:tree-eg-depth2} as an example, where we have three binary variables $X,Y,Z\in(0,1)$. This tree has two paths that terminate in a $D=1$ leaf, and they correspond to the rules ($Z=0$ AND $Y=1$) and ($Z=0$ AND $X=1$). The corresponding rule set is simply the union of these rules. The first limitation is immediately obvious: Every rule must include a condition with the variable corresponding to the root node, in this case $Z$, because every path must start with that node. 
Since sub-trees are also trees, this logic extends recursively for all lower nodes; all paths that go down the left subtree must also include a condition on that tree's root node, in this case $Y$, and similarly for the right subtree.

The second limitation is more subtle. At each node, we can only have one branch (and thus one child node) corresponding to each value of that node's variable. Thus, if you have two rules that overlap in conditions, you will need a much deeper tree to represent their union. Take for example ($Z=0$ AND $Y=1$ AND $X=1$) OR ($Z=0$ AND $Y=1$ AND $W=1$). You cannot simply create a tree where the root is $Z$, one of its children is $Y$, and then its two children are $X$ and $W$, because at the node $Y$ there is only one branch corresponding to $Y=1$ that can represent either $X=1$ or $W=1$ but not both. You also cannot use any of the other subtrees to represent the remaining rule, since they will necessarily include the condition $Z=1$ or $Y=0$. Instead, we would need to rearrange the tree to begin with a condition that is neither $Z$ nor $Y$ so that we can make use of the other subtree it produces and include $Z$ and $Y$ later on. For example, we could start with the node $X$, and have one path that leads to $Z$ then $Y$ to encode the first rule. To encode the second rule, we would need to have all three variables $Z$, $Y$, and $W$ be descendants of $X$, making our tree depth four. That is, in order to represent a rule set with two rules of length three each, we need a tree of depth four. Note that a tree of depth four can have as many as $2^4$ leaves, half of which can be $D=1$, and so in the worst case such a tree could represent a rule set with eight rules each of length four.

We use Figure~\ref{fg:complexity} to represent the comparison between the capacities of rule sets and trees to represent decision rules of the same complexity. Rule sets are much more flexible, and so we can be much more restrictive about the size of the feasible class while retaining the capacity to represent adequately complex decision rules. Trees have a much more restrictive structure, and so to attain the same representation, we would need to grow the size of the feasible class to the point that it also includes highly uninterpretable solutions.

\subsection{Pareto Efficient Frontier}
Within our set of feasible, interpretable solutions, we focus on two desiderata---treatment effect size and subgroup size---and the trade-off between them. At one extreme, we could identify the single observation with the highest treatment effect. As we increase the subgroup size by adding more observations, we necessarily will add observations with lower effect sizes. 
Any reasonable solution $A$ should satisfy a Pareto efficiency condition: There does not exist another feasible rule set that dominates $A$ in both support and effect size, $\nexists A'\in\mc{A}$ s.t. $\supp(A) < \supp(A')$ and $\hat\tau(A) < \hat\tau(A')$. 
A collection of such rule sets forms a Pareto efficient frontier of rule sets, or a Pareto front.
Note that this is identical to the concept of an ``efficient treatment-frontier'' from \citet{wang2022causal}.
In the next section, we introduce an objective function that trades off between support and effect size and discuss how tuning its hyperparameter will yield different points on the Pareto front.



\section{An Algorithm for Subgroup Discovery}

We introduce a variant of the simulated annealing algorithm for the task of subgroup discovery. Simulated annealing is a popular heuristic in discrete optimization due to its theoretical proved convergence and empirically observed performance. It balances exploration with exploitation, favoring the former earlier and the latter later in the chain, to avoid becoming trapped at local optima. We provide a sketch of our algorithm in Algorithm \ref{alg:search}.

\subsection{Setup}

Assume we have a set of estimates of the individual treatment effects (ITE) $\{\hat\tau_i\}_{i=1}^N$ for each observation.
It suffices, for example, to have a fitted model $\hat\tau(x)$ for the CATE, in which case we can then use $\{\hat\tau(x_i)\}_{i=1}^{N}$ as the set of ITE estimates.
$X\in\mc{X}$ denotes the covariates, which we assume are discrete. 
We use $\mc{A}$ to denote the collection of all (feasible) rule sets. With some abuse and overload of notation, we use $A(\cdot)$ as an indicator function for membership in the set $A$, and we use $\tau(A)=\E[\tau(x)|A(x)=1]$ to denote the CATE for observations in $A$. The general goal is to find a rule set $A$ such that $\tau(A)>c$ for some threshold $c$ such as the ATE or zero.


\subsection{Objective Function}
We propose a multi-objective objective function that trades off subgroup size and subgroup treatment effect size,
\begin{equation}\label{eq:obfn}
	F(A, X, \hat\tau; \alpha) = \Bigg(\frac{\supp(A)}{N}\Bigg)^\alpha \frac{\sum_{i \in A} \itehat - \min_i\itehat}{\max_i\itehat}, 
\end{equation}
where $\supp(A)=|\{i:i\in A\}|$ denotes the support of a rule set $A$ (i.e., the number of observations satisfying the rule set) and $N$ is the total number of observations. $\frac{\supp(A)}{N}$ is thus the proportion of observations satisfying the rule set $A$. 

$\alpha \geq 0$ is a hyperparameter that controls the amount of weight we put on the size of a subgroup. When $\alpha=0$, we place no weight on subgroup size, since the objective function is constant with respect to subgroup size, and subgroup size becomes increasingly important as $\alpha$ grows. We see this if we examine the partial derivatives of $F$ for various values of $\alpha$. If we denote $p_{\supp}=\frac{\supp(A)}{N}$ and $p_{\itehat}=$  $\alpha p^{\alpha-1}q$ and $p^\alpha=\frac{\sum_{i \in A} \itehat - \min_i\itehat}{\max_i\itehat}$, then $\partial F/\partial p_{\supp} = \alpha p_{\supp}^{\alpha-1}p_{\itehat}$ and $\partial F/\partial p_{\itehat}=p_{\supp}^\alpha$.
If we take the ratio of these two, we get $\alpha p_{\itehat}/p_{\supp}$. Larger values indicate that improvements in $p_{\supp}$ are more important relative to improvements in $p_{\itehat}$ than smaller values. Clearly the function is monotonically increasing in $\alpha$. 

This objective function also has the desirable property that any maximizer must be Pareto optimal. That is, if a rule set $A^*=\argmax_{A\in \mc{A}} F(A, X, \itehat; \alpha)$ for any $\alpha > 0$, then there does not exist another rule set $A$ that dominates $A^*$ in both support and effect size, $\supp(A) > \supp(A^*)$ and $\hat\tau(A) > \hat\tau(A^*)$. This is easy to see, since $F$ is monotonically increasing in both $p_{\supp}$ and $p_{\itehat}$.\footnote{It is not, however, conversely guaranteed any Pareto optimal rule set must also be a solution. An easy counterexample is for any ruleset where $\hat\tau(A)=0$: In this case, $F(A)=0$ regardless of the support of $A$.
}

Although we do not do so in the current paper, the objective function can be further customized to include other objectives of interest. A regularization term can enhance interpretability further and help prevent overfitting. We observe that overfitting is especially a concern when searching for for rule sets with high effects and small support, where (1) higher order interactions may exist in the training set but not in the test set or the population more broadly and (2) there may be rules with very low support that would not be worth ``spending" complexity on when more weight is placed on subgroup size. In line with this observation, it could be sensible to weight how much we regularize a term based on the support of a rule. If harm prevention is especially important, we can decompose the term on effect size to disproportionately punish rule sets for negative treatment effects. Fairness is also a common objective, various measures can also be incorporated. The primary difficulty with adding additional terms is that parameter and result selection tuning becomes increasingly difficult, and the strategies we recommend in this paper do not necessarily extend to other object functions.


\paragraph{Interpretability constraints.} In lieu of regularization, we pursue a constrained optimization strategy for simplicity. In particular, we constrain the maximum length of each rule $L_{max}$ and the total complexity of a rule set $C_{max}$, where a length of a rule is defined as the number of conditions in it, and the complexity of a rule set is the sum of the lengths of its constituent rules. Placing a maximum length on rules also serves a key function in limiting the search space.



\subsection{Neighbor Selection}
We define the neighbor of a rule set in the same way as Wang et al, but we modify their approach for selecting neighbors in several ways to fit our setting. 
A rule set $A$ is a neighbor of a rule set $A'$ if and only if $A$ differs from $A'$ by at most one rule. That is, $A$ can be $A'\setminus a_l$ (remove a rule), $A'\cup a_l$ (add a rule), or $(A'\setminus a_l)\cup a_l$ (replace a rule). 
We propose a dynamic neighbor selection strategy that makes larger changes to the solution early on and smaller changes later on.
We define the ``fine-gained neighborhood'' of $A'$ as the set of rule sets that differ by at most one condition. 
At each step, when selecting a neighbor, we choose to pick from the fine-grained neighborhood with probability $p_{fg}$ and the general neighborhood otherwise. We choose $p_{fg}$ to increase over time, so that the algorithm makes smaller changes later in the chain. We use the logistic function $\frac{1}{1+\exp(-fg_{scale}(t-fg_{switch}Niter)/Niter)}$, where $t$ is the current iteration and $Niter$ is the maximum iteration, in our examples. This choice is such that the probability of choosing from a fine-grained neighborhood eclipses the probability of choosing from the general neighborhood after $fg_{switch}Niter$ of the iterations, with the steepness of the transition being controlled by $fg_{scale}$.

\subsection{Search Procedure}
Once we choose the type of neighborhood, we randomly choose a type of action. In the general neighborhood, the possible actions are to add a rule, remove a rule, or replace a rule, denoted \add, \cut, and \replace. In the fine-grained neighborhood, we first choose a rule that is in the rule set, then choose to add a condition to it, remove a condition from it, or replace a condition in it, denoted \addc, \cutc, \replacec.
Conditional on the type of neighborhood and type of action, we pursue an $\epsilon$-greedy strategy for selecting neighbors: With some probability $q$, we pick the rule or condition uniformly at random (exploration), and we pick the rule or condition that maximizes the objective function otherwise (exploitation). 
We detail these actions here.
\begin{itemize}
	\item \add: 
	\begin{enumerate}
		\item Select a rule $a$.
		\begin{itemize}
			\item uniformly at random with probability $q$, or
			\item to maximize the objective function, $a=\argmax_{a'} A_t\cup a'$.
		\end{itemize}
		\item Propose $A_{t+1}=A_t\cup a$.
	\end{enumerate}
	\item \cut: 
	\begin{enumerate}
		\item Select a rule $a\in A$.
		\begin{itemize}
			\item uniformly at random with probability $q$, or
			\item to maximize the objective function, $a=\argmax_{a'} A_t\setminus a'$.
		\end{itemize}
		\item Propose $A_t\setminus a$.
	\end{enumerate}
	\item \replace: \cut, then \add.
	\item \addc: 
	\begin{enumerate}
		\item Select a rule $a \in A$.
		\item Select a condition $z\notin C$, where $a=\bigcap_{z'\in C} z'$
		\begin{itemize}
			\item uniformly at random with probability $q$, or
			\item to maximize the objective function, $z=\argmax_{z'} (A_t\setminus a)\cup (a\cap z') $.
		\end{itemize}
		\item Propose $(A_t\setminus a)\cup (a\cap z)$.
	\end{enumerate}
	\item \cutc: 
	\begin{enumerate}
		\item Select a rule $a\in A$.
		\item Select a condition $z \in C$, where $a=\bigcap_{z'\in C} z'$
		\begin{itemize}
			\item uniformly at random with probability $q$, or
			\item to maximize the objective function, $z=\argmax_{z'} (A_t\setminus a)\cup (\bigcap_{c\in (C\setminus z')}c)$
		\end{itemize}
		\item Propose $(A_t\setminus a)\cup (\bigcap_{c\in (C\setminus z)} c)$.
	\end{enumerate}
	\item \replacec: \cutc, then \addc.
\end{itemize}

We accept a proposal with probability $\min\{1, \exp(\frac{F(proposal)-F(A_t)}{T_t})\}$, where $T_t=T_0 \eta^{t}$ for some $\eta$ close to 1 that may depend on $Niter$. We propose calibrating $T_0$ prior to the start of the search by randomly proposing solutions, calculating the change in the objective function it would induce $\Delta$, and setting $T_0$ to be slightly larger than a ``typical'' $\Delta$. For example, we could use the 75th quantile of $\Delta$'s, or, as we do in our examples, the mean of $\Delta$'s scaled up by a small factor. This strategy will produce acceptance probabilities that are almost always near 1 early on and close to 0 later in the chain unless the proposal improves the objective function.

\begin{algorithm} 
\caption{Rule set search} 
\label{alg:search} 
\begin{algorithmic} 
    \REQUIRE $\{\itehat, X_i\}_{i=1}^N, \alpha, L_{max}, C_{max}, NRules, NIter, p_{fg}, q$
    \STATE Mine for rules of length at most $L_{max}$
    \STATE Cull to have $NRules$ rules
    \STATE Randomly sample $A_0$
    \STATE Calibrate cooling schedule parameter $T_0$
    \FOR{$t=0$ to $NIter$}
    	\STATE sample $u\sim Uniform(0,1)$
    	\IF{$u < p_{fg}$}
    		\STATE Uniformly sample $a$ from $A_t$ 
	    	\IF{length($a_i$)$=L_{max}$}
	    		\STATE \texttt{ACTION} $\leftarrow\begin{cases}\texttt{CUT}_{cond} \text{ w.p. } 1/2 \\
	    			\texttt{REPLACE}_{cond} \text{ w.p. } 1/2\end{cases}$
	    	\ELSIF{length$(a_i)=1$}
	    		\STATE $\texttt{ACTION}\leftarrow\begin{cases}
	    		\texttt{ADD}_{cond} \text{ w.p. } 1/2 \\ 
	    		\texttt{REPLACE}_{cond} \text{ w.p. } 1/2\end{cases}$
	    	\ELSE
	    		\STATE \texttt{ACTION}$\leftarrow\begin{cases}
	    		\texttt{ADD}_{cond} \text{ with probability } 1/3 \\
	    		\texttt{CUT}_{cond} \text{ with probability } 1/3 \\ 
	    		\texttt{REPLACE}_{cond} \text{ w.p. } 1/3\end{cases}$
	    	\ENDIF
	    	\STATE $a' \leftarrow$ \texttt{ACTION}($a$)
	    	\STATE $A_{t+1}\leftarrow\begin{cases}
	    		(A_{t} \setminus a)\cup a' \text{ w.p. } \pi \\
	    		A_{t} \text{ w.p. } \pi
	    	\end{cases} $
		\ELSE
	        \IF{$\sum_{a_i \in A_t}$length($a_i$) $=C_{max}$}
		    	\STATE \texttt{ACTION} $\leftarrow\begin{cases}
		    		\texttt{CUT} \text{ w.p. } 1/2 \\ 
		    		\texttt{REPLACE} \text{ w.p. } 1/2\end{cases}$
			\ELSIF{length($A_t$)=1}
				\STATE $\texttt{ACTION}\leftarrow\begin{cases}
		    		\texttt{ADD} \text{ w.p. } 1/2 \\ 
		    		\texttt{REPLACE} \text{ w.p. } 1/2\end{cases}$
			\ELSE
		    	\STATE \texttt{ACTION}$\leftarrow\begin{cases}
		    		\texttt{ADD} \text{ w.p. } 1/3 \\
		    		\texttt{CUT} \text{ w.p. } 1/3 \\ 
		    		\texttt{REPLACE} \text{ w.p. } 1/3\end{cases}$
	    	\ENDIF
	    	\STATE $A_{t+1}\leftarrow\begin{cases}
	    		 \texttt{ACTION}(A_t) \text{ w.p. } \pi \\
	    		A_{t} \text{ w.p. } \pi
	    	\end{cases} $	
    	\ENDIF
    \ENDFOR
    \RETURN $A_{max}$
\end{algorithmic}
\end{algorithm}








\subsection{Inference}
We rely on sample splitting to conduct inference, similar to the principle behind ``honest trees'' in \citet{wager-athey_2018_forest}. We consider two types of hypothesis tests, which depend on the type of threshold $c$ is.

\paragraph{Fixed threshold.}
If $c$ is a fixed number then our hypothesis test is simple. For example, if $c=0$, we might test $H_0: \tau(A) = 0$ (or potentially $H_0: \tau(A) < 0$). We can then simply estimate $\hat\tau(A)$ in the hold out sample and conduct a standard hypothesis test. 
In our empirical example, since we have a randomized experiment, we simply take the difference in means between $Y(1)$ and $Y(0)$ for observations covered by $A$.


\paragraph{Comparing groups.}
If $c$ is itself an unknown quantity, then we will need to account for additional uncertainty. For example, if $c$ is the ATE $\tau$, then $H_0: \tau(A) = \tau$ (or $H_0: \tau(A) < \tau$). 

A more general statement of the problem is as follows. We are interested in testing the hypothesis $H_0: \Delta = 0$, where $\Delta=\hat\tau_A-\hat\tau_B$, using the statistic $\hat{\Delta}=\hat{\tau}_A - \hat{\tau}_B$, where $\hat\tau_A$ and $\hat\tau_B$ are themselves statistics. The formula for uncertainty propagation for a linear function $f(\bf{x})=\bf{ax}$, where $\bf{a}\in\R^m$ and $\bf{x}$ is a random vector with covariance matrix $\bf{\Sigma_x}$, is $\sigma_f^2 = \bf{a\Sigma_x a}^T.$ For $f=\hat\Delta$, $\bf{a}=[1, -1]$ and $\bf{x}=[\hat\tau_A, \hat\tau_B]$, so $\sigma_{\hat\Delta}^2 = \sigma_A^2+\sigma_B^2-2\sigma_{AB}$, where $\sigma_A^2$ and $\sigma_B^2$ are the variances of $\hat\tau_A$ and $\hat\tau_B$ and $\sigma_{AB}$ is their covariance. 
Often bootstrap may be used when closed-form estimators are not known, although this comes with the usual caveats that bootstrap does not work for all estimators (e.g., LASSO or matching) and still relies on asymptotics.





\paragraph{Power calculations.} In our empirical example, we conduct power calculations using the estimated values of $\tau(A)$ from the training set and a sample size based on the size of the test set and the support of $A$ in the training set. We use a two-sided hypothesis test to be conservative, although the ranking of power should in general be preserved if switching to a one-sided test.

\section{Simulation}

We use two simulated examples to demonstrate the capabilities and limitations of our method. First, we use a simple data generating process (DGP) where variables are discrete and the treatment effects are determined using rule-based logic. We show that our method recovers all Pareto optimal points when the front is concave and some but not all Pareto optimal points when the curve is non-concave. We compare this with a version of our algorithm that uses the linear objective function, which is unable to recover any interior points in the non-concave setting.

Second, we use a more adversarial DGP where variables are continuous and treatment effects are determined by continuous functions. We show that even in this setting, our method is able to recover a reasonable characterization of the data

\subsection{A Simple Simulated Example}
In the simple setting, we have $N=1000$ examples and $J=10$ binary variables. We generate the data as follows:
\begin{enumerate}
	\item Set $\tau_i=0$ for all observations;
	\item For $l=1,2,3$:
	\begin{itemize}
		\item[-] Set $\tau_i=\mu_i$ for all $i\in a_l$,
	\end{itemize}
	where $a_1=A$, $a_2=B\cap C$, and $a_3=D\cap E\cap F$.
\end{enumerate}

To generate a concave front, we use $\mu=(4.5, 6.5, 7)$, and to generate a convex front, we use $\mu=(1,5,10)$.

\begin{figure}[h!]
\caption{Simulated Example: Discrete Setting}
\label{fg:simple-sim}
\hfill
\begin{subfigure}[t]{0.475\textwidth}
         \includegraphics[width=\textwidth]{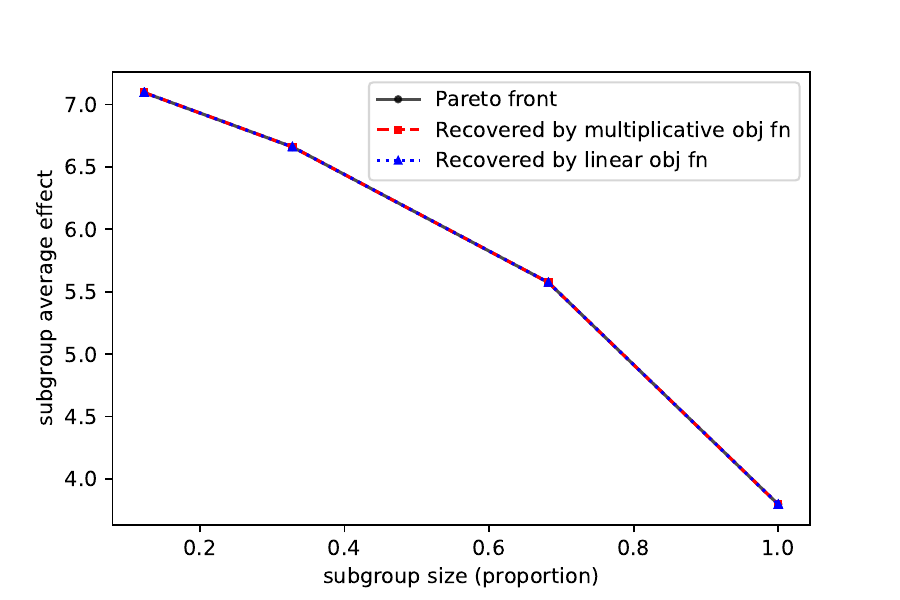}
         \caption{Concave Pareto Front}
         \label{fg:concave}
     \end{subfigure}
     \hfill
     \begin{subfigure}[t]{0.475\textwidth}
         \includegraphics[width=\textwidth]{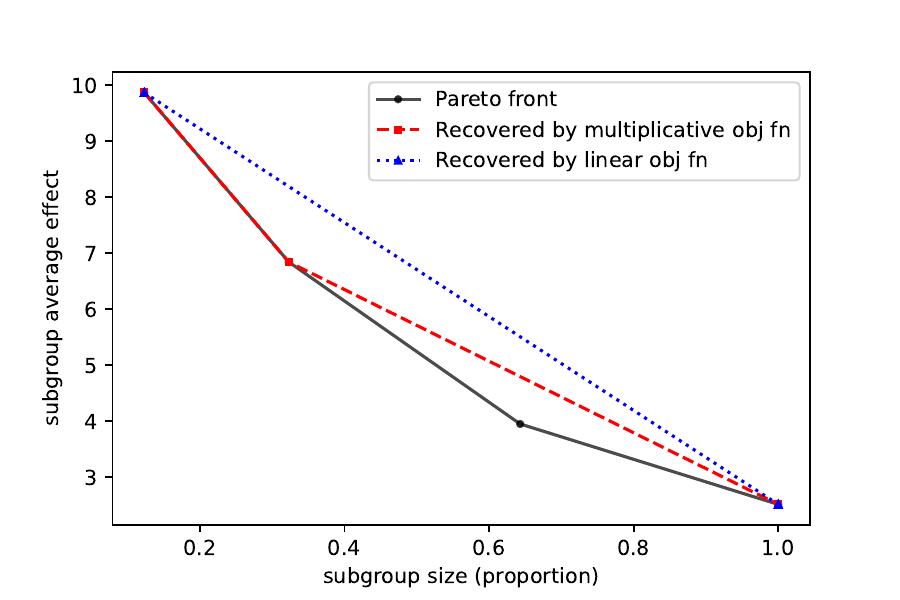}
         \caption{Convex Pareto Front}
         \label{fg:convex}
     \end{subfigure}
     
     \footnotesize{\textbf{Note:} Comparison of the Pareto optimal points (black solid line, circle marker) with those that the multiplicative objective function can recover (red dashed line, square marker) and that the linear objective function can recover (blue dotted line, triangle marker). Both objective functions are able to recover all Pareto optimal points when the front is concave (left). When the front is convex (right), the linear objective function is unable to recover any interior points, and the multiplicative objective function is able to partially recover the interior points.}
\end{figure}

In Figure~\ref{fg:simple-sim}, we display the frontiers generated by our method when using our recommended multiplicative objective function (red dashed line, square marker) and a linear objective function (blue dotted line, triangle marker), along with the true frontier of Pareto optimal points based on our DGP ((black solid line, circle marker). 
In Figure~\ref{fg:concave}, we see that our method recovers all Pareto optimal points when using either objective function. In Figure~\ref{fg:convex}, however, we see that the linear objective function is incapable of recovering any interior point when the front is convex.\footnote{In the section on multi-objective optimization, where we stuck to convention and assumed the problem was a minimization task, we noted it could always recover points on convex fronts but not concave ones. Since our goal is maximization here, convexity and concavity are reversed.}
The multiplicative objective function, on the other hand, can recover part of but not all of the solutions lying on the convex curve.

We make a secondary observation that it does not seem like our method is prone to overfitting, even without randomization; it did not recover solutions that were simply artifacts of this particular randomization and instead only recovered ``true'' treatment assignment rules. We present further evidence of this phenomenon in our second simulated example and our empirical example.

\subsection{An Adversarial Example}

We use a larger sample size of $N=10,000$. We have 10 continuous variables $X_1,\dots,X_{10}$, where $X_j\sim\uniform(0,1)$ for all $j$. We set $\tau(x) \propto (X_1 + X_2 + 1)(X_4 + X_5 + 1)(X_7 + X_8 +1)$. This results in a variety of interactions of various degrees: There are six terms with no interaction, 12 two-way interactions, and eight three-way interactions. $\tau(x)$ is also monotonic in all dimensions, making it easier to validate the directionality of the conditions that appear in our rule set.
We discretize variables based on quantiles, then create indicators for a sequence of growing sets. In particular, for each $X$, we create three binary indicators for $X < q_{25}$, $X < q_{50}$, and $X < q_{75}$, where $q_p$ is the $p$th quantile of $X$.

As a face validity check, our method seems to perform well. First, amongst the 69 variable combinations that we identify, 58 of them (84.1\%) are ``real'' interactions. E.g., a rule containing conditions on $X_1$ and $X_4$ counts in this category, since the term $X_1X_4$ appears in the expansion of $\tau(x)$. Second, amongst the 170 conditions that appear (including repeated appearances), 161 (94.7\%) were $\geq$ conditions (with the remaining being $<$ conditions).

\begin{figure}[h!]
\caption{Simulated Example: Continuous Setting}
	\begin{center}
	\includegraphics[width=.7\textwidth]{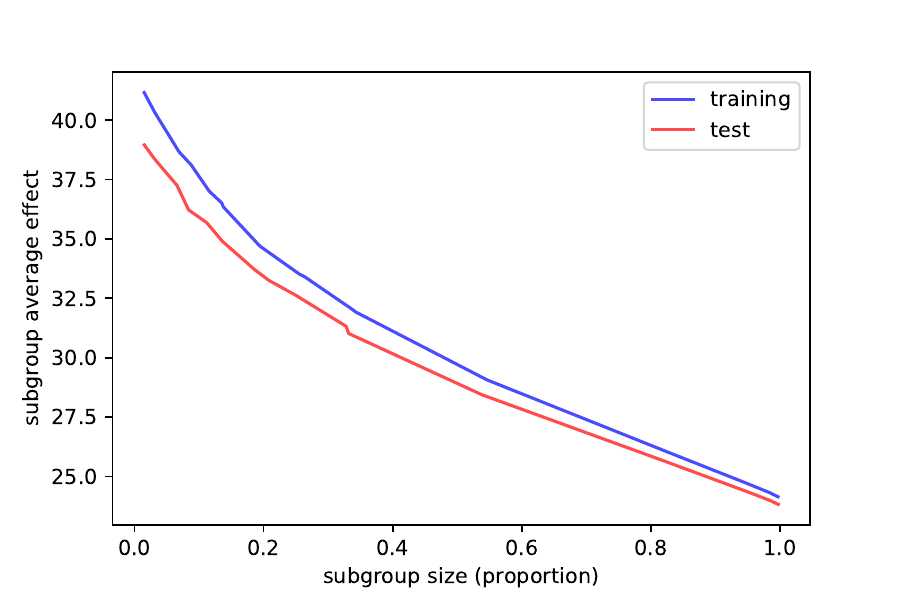}
	\end{center}
	\label{fg:sim-cont}
	\footnotesize{\textbf{Note:} Comparison of the training and test set performance of Pareto optimal points obtained using our method on a simulated example with continuous covariates.}
\end{figure}

Figure~\ref{fg:sim-cont} plots the subgroup average treatment effect against the support rate of each rule set on the Pareto front we identify. We use a 70/30 training/test split, and the blue line shows the estimates on the training set, while the red line shows the estimates on the test sets. We see some indication of overfitting, but the test set performance recovers as the support grows.

\section{Empirical Example}
We use data on the effectiveness of the youth job training program Job Corps to demonstrate our method. In 1993, the Department of Labor commissioned the National Job Corps Study, which randomly assigned 81,000 eligible applicants aged 16 to 24 to either treatment, in which case they were eligible to enroll, or control, in which case they were not eligible. Surveys were conducted in both groups in the four years after random assignment, and administrative data on earnings are also available \citep{schochet2008does}.

We use causal forest \citep{wager-athey_2018_forest} to estimate the CATE of assignment to treatment on self-reported weekly earnings three years after assignment using the variables age (discretized by quantile), race (binary indicators for white, black, and hispanic), native language (binary indicators for English and Spanish), frequency of receipt of welfare by respondent when growing up, marriage, children (binary indicator), education (any degree, less than high school, and at least high school), employment in the previous year or currently (binary indicator for at least one being true), any education or training in the past year (binary indicator), current receipt of welfare (binary indicator), personal income (discrete ranges), arrest history (binary indicator), guilty verdict (binary indicator for ever convicted, pled guilty, or adjudged), and served time in jail (binary indicator).\footnote{For the variables age, receipt of welfare growing up, and personal income, we use increasing series of sets, e.g., $age \leq age_{25}$, $age \leq age_{50}$, and $age \leq age_{75}$, where $age_q$ refers to the $q$th quantile of $age$. Receipt of welfare growing up was reported as ``never,'' ``occasionally,'' ``half of the time,'' and ``always.'' Data on income was reported as ranges of $<\$3,000$, \$3,000 to \$6,000, \$6,000 to \$9,000, and $>\$9,000$.}
After subsetting to respondents with complete data for these variables, we are left with 7,830 observations. 

Once we have our estimates of the CATE, we split our data into training and test sets along a 70/30 split. We vary $\alpha$ to produce a number of solutions and retain the non-redundant and non-dominated ones. We also obtain results using the \texttt{policytree} package \citep{sverdrup2020policytree}.

\begin{figure}[h!]
\caption{Selected Treatment Rules}
	\begin{center}
	\includegraphics[width=.9\textwidth]{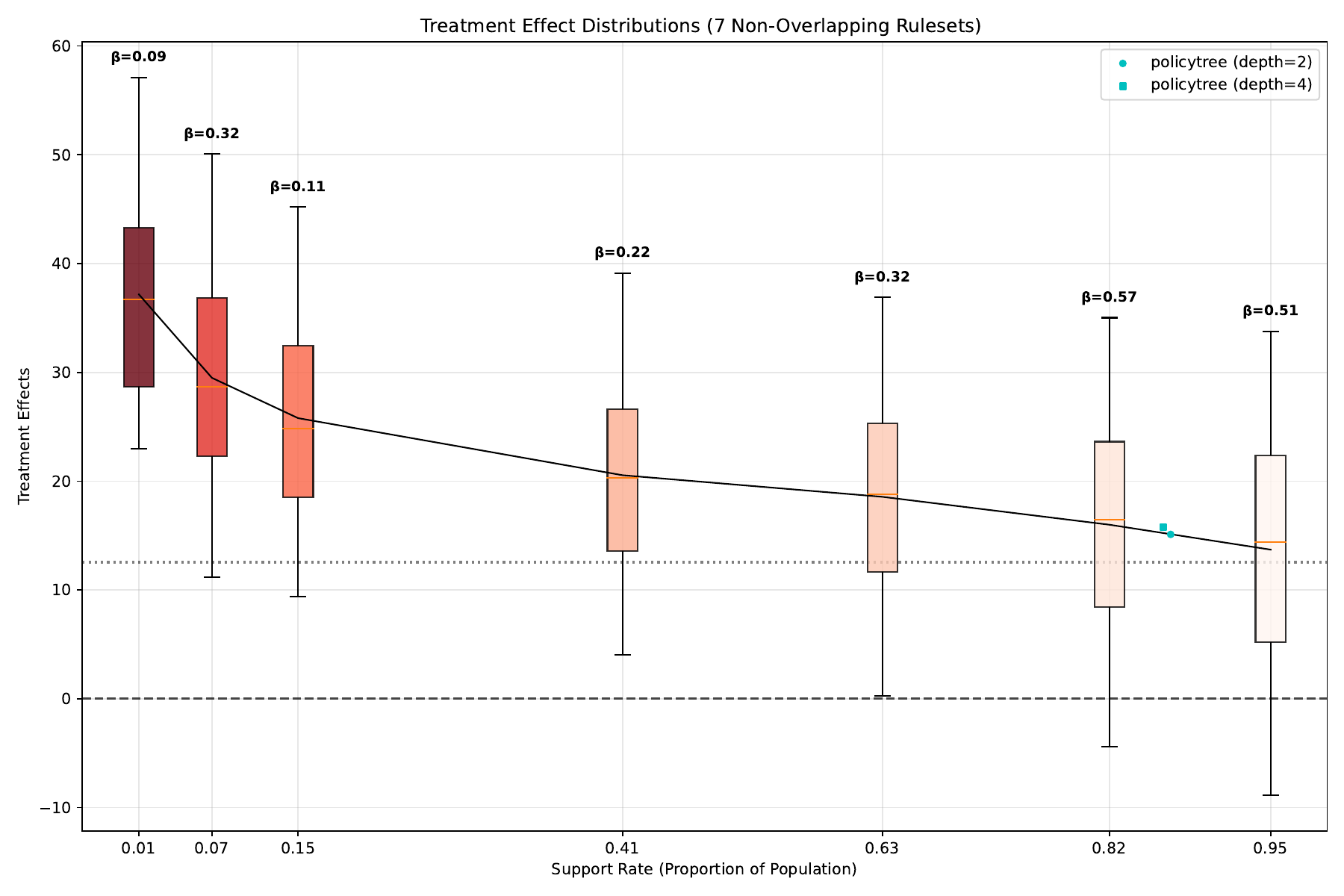}
	\end{center}
	\label{fg:box}
	\footnotesize{\textbf{Note:} Each box plot represents a rule set on the Pareto front. Effect sizes and support are both calculated using the training set. Dotted line represents the overall average treatment effect.} 
\end{figure}

In Figure~\ref{fg:box}, we display a subset of the selected rules, removing ones that were overlapping for visual clarity. We plot distribution of in-sample (training) subgroup treatment effects for each rule set against their support rate (support as a proportion of the sample size). A blue dot and square mark the trees obtained using \texttt{policytree} with depths set at two and four.

Our first observation is that our method is significantly more flexible than \texttt{policytree}. \texttt{policytree} is designed only to maximize only overall welfare and can not easily trade off between group size and effect size.
While in theory the proposal by \citet{athey-wager_2021_policy} is highly flexible, it is not clear in practice how we would encode most desiderata. 
The authors' general strategy is to first define a class of policies $\Pi$ that satisfies our preferences, then to find the optimal solution $\pi\in\Pi$. The authors motivate alludes to the ability of $\Pi$ to encode budget, fairness, and functional form constraints \citep{zhou2023offline}, they are only clear on how to generate $\Pi$ respecting functional form constraints.
It is not feasible, for example, to enumerate the entire class of policies and then subset to the ones satisfying our constraints, as the size of this class can grow exponentially.

We also calculate the power of a two-sample, two-sided $t$-test for $\E[Y(1)|A]=\E[Y(0)|A]$ for each rule set $A$, where we assume the mean and standard deviation of $Y(1)$ and $Y(0)$ are their sample averages amongst covered observations in the training set (which is only appropriate in this case since treatment is assigned randomly), and $N_1$ and $N_0$ are estimated as the product of the support rate and the size of the test set. As expected, rule sets with minimal coverage have low power despite higher effect sizes, since the gain in effect size (at most double) is small relative to the precipitous decline in support. The rule sets with maximal power are those with relative large support (over 0.8) but still above average (dotted line) and well-above zero CATEs.

This observation suggests a potential use case for our method: To maximize the probability of identifying some treatment rule with positive average treatment effects. These power calculations give us a criteria for selecting a rule set from the Pareto front.
Depending on the context, we can use other hypotheses (e.g., above some threshold for which it is worth treating an individual, or the probability of a negative effect is below some threshold).

\begin{figure}[h!]
\caption{Training and Test Set Performance of Selected Rule Sets}
	\begin{center}
	\includegraphics[width=.9\textwidth]{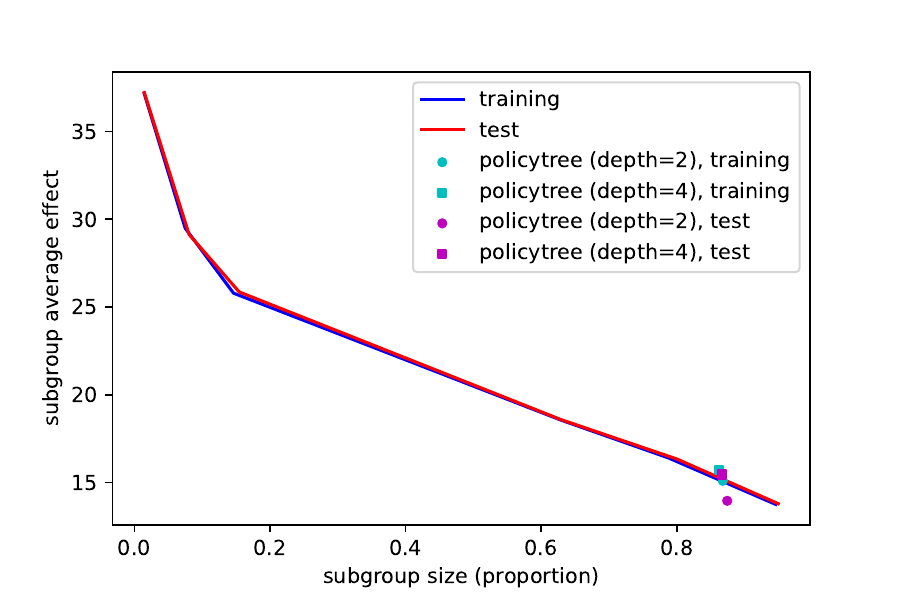}
	\end{center}
	\label{fg:trainingtest}
	\footnotesize{\textbf{Note:} Comparison of the training and test set performance of Pareto optimal points obtained using our method and \texttt{policytree}.}
\end{figure}

In Figure~\ref{fg:trainingtest}, we plot the training and test set performance of the selected rule sets together. In this example, it seems that there is not noticeable overfitting: The training and test curves are very similar, and the only notable drop in test set performance is by the depth two tree.

\section{Conclusion}
This paper introduces a novel method for discovering interpretable, rule-based treatment assignment rules.
We use a multiplicative objective function treatment assignment rule that trades-off between treatment effect size and subgroup size and vary the hyperparameter controlling this trade-off to find a frontier of Pareto optimal rule sets.
Our objective is a middle ground between the linear scalarizations that are easier to optimize but cannot represent interior solutions on convex fronts and the hypervolume scalarization, which can represent any Pareto optimal solution but is a more challenging problem to solve.
We show that the multiplicative scalarization can capture some Pareto optimal points on convex fronts that a linear objective function misses.
  
We propose a search procedure based on simulated annealing that balances exploration with exploitation. We use complexity constraints to ensure interpretability. We use an adaptive neighborhood selection criteria that is more likely to make smaller changes later in the chain. Valid inference on the treatment effect of any given solution is possible with sample splitting, and we suggest power calculations based on estimates from the training sample as a potential criteria for selecting rule sets from the frontier.

In two simulated examples, we demonstrate that our method recovers reasonable rule sets, both when the DGP is favorable and when it is adversarial. When variables are discrete and treatment effects are assigned using rule-based logic, our method recovers all points on concave Pareto fronts and most points on convex ones. When variables are continuous and the CATE is a smooth function, our rule set recovers rules with the correct directionality and interactions with high probability.
Lastly, in an empirical example on the effects of the job training program Job Corps, we show that our method is more flexible than methods that optimize only overall welfare. Evaluations on the test set suggest our method is not prone to overfitting, even without regularization.

In future work, we hope to develop computational methods for solving the more robust but difficult hypervolume scalarization. We also acknowledge that stability is a key challenge to rule-based methods \citep[e.g.,][]{chiu2023bayesian} and hope to make progress in increasing the stability of results.
Despite these limitations, we argue our method has the ability to discover interpretable treatment rules that capture complex interactions. We hope that our method will benefit applications where interpretability and transparency are key, including medicine, government policy, and industry.

\onehalfspacing
\bibliographystyle{apsr}
\bibliography{hte-fontier_120625.bib}

\end{document}